%% file: main.tex
\documentclass[10pt,twocolumn,letterpaper]{article}

\usepackage{cvpr}              

\input{preamble}

\definecolor{cvprblue}{rgb}{0.21,0.49,0.74}
\usepackage[pagebackref,breaklinks,colorlinks,allcolors=cvprblue]{hyperref}

\def\paperID{16895} 
\def\confName{CVPR}
\def\confYear{2025}

\title{Video-DPRP: A Differentially Private Approach for Visual Privacy-Preserving Video Human Activity Recognition}

\author{Allassan Tchangmena A Nken$^{1}$
~~Susan Mckeever$^{2}$
~~Peter Corcoran$^{1}$
~~Ihsan Ullah$^{1}$\\
$^{1}$ University of Galway, Ireland,~~ $^{2}$ Technological University Dublin, Ireland
}

\begin{document}
\maketitle
\input{sec/0_abstract}    
\input{sec/1_intro}
\input{sec/2_formatting}
\input{sec/3_finalcopy}
\input{sec/4_experiment}
{
    \small
    \bibliographystyle{ieeenat_fullname}
    \bibliography{main}
}


\end{document}

%% file: preamble.tex
\usepackage{graphicx}
\usepackage{amsmath}
\usepackage{amssymb}
\usepackage{booktabs}
\usepackage{multirow}
\usepackage{amsthm}
\usepackage{algorithm}
\usepackage{algpseudocode}
\usepackage{colortbl}  
\usepackage{xcolor}
\usepackage{tabularx}

\newtheorem{theorem}{Theorem}
\newtheorem{lemma}[theorem]{Lemma}
\newtheorem{definition}[theorem]{Definition}

%% file: sec/0_abstract.tex
\begin{abstract}
 Considerable effort has been made in privacy-preserving video human activity recognition (HAR). Two primary approaches to ensure privacy preservation in Video HAR are differential privacy (DP) and visual privacy. Techniques enforcing DP during training provide strong theoretical privacy guarantees but offer limited capabilities for visual privacy assessment. 
Conversely methods, such as low-resolution transformations, data obfuscation and adversarial networks, emphasize visual privacy but lack clear theoretical privacy assurances. In this work, we focus on two main objectives: 
(1) leveraging DP properties to develop a model-free approach for visual privacy in videos and (2) evaluating our proposed technique using both differential privacy  and visual privacy assessments on HAR tasks. 
To achieve goal (1), we introduce \textbf{Video-DPRP}: a \textbf{Video}-sample-wise \textbf{D}ifferentially \textbf{P}rivate \textbf{R}andom \textbf{P}rojection framework for privacy-preserved video reconstruction for HAR. 
By using random projections, noise matrices and right singular vectors derived from the singular value decomposition of videos, Video-DPRP reconstructs DP videos using privacy parameters ($\epsilon,\delta$) while enabling visual privacy assessment. For goal (2), using UCF101 and HMDB51 datasets, we compare Video-DPRP's performance on activity recognition with traditional DP methods, and state-of-the-art (SOTA) visual privacy-preserving techniques. Additionally, we assess its effectiveness in preserving privacy-related attributes such as facial features, gender, and skin color, using the PA-HMDB and VISPR datasets. Video-DPRP combines privacy-preservation from both a DP and visual privacy perspective unlike SOTA methods that typically address only one of these aspects.
\end{abstract}

%% file: sec/1_intro.tex
\section{Introduction}
\label{sec:intro}
Privacy preservation is a critical research challenge in the field of video-based human activity recognition (HAR) and video analysis. Video HAR systems are increasingly used in settings like healthcare monitoring, smart homes and security \cite{jalal2012depth,zhou2020deep,cristina2024audio,shojaei2018video}. However, these systems often capture sensitive personal information, creating a strong need for privacy measures to protect individuals' identities and personal activities from misuse or unauthorized access. 

\begin{figure*}[t]
\includegraphics[width=1\linewidth]{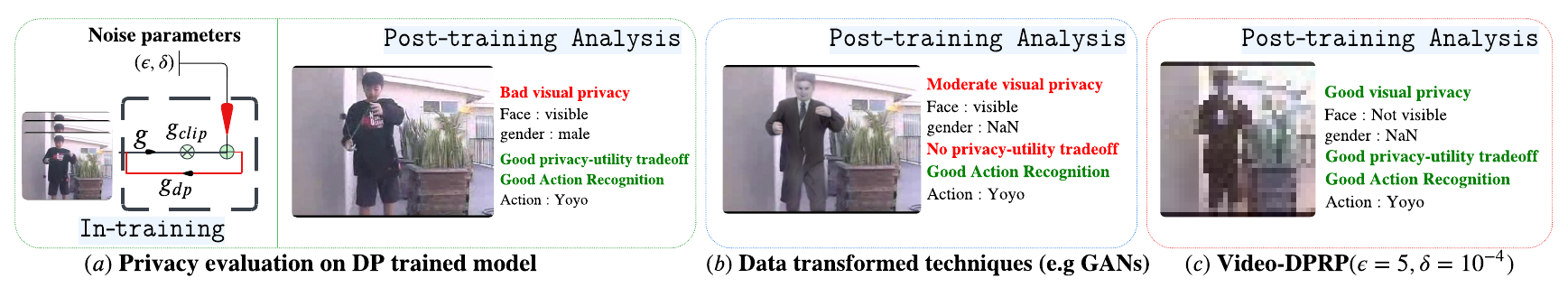}
\caption{In $(a)$, privacy is ensured during training (in-training) using differential privacy (DP), but not directly on the video itself. As a result visual privacy cannot be assessed. In $(b)$, the video is transformed prior to training using either obfuscation methods or adversarial approaches, but the privacy-utility trade-off cannot be quantify as clearly as in DP. In $(c)$ (ours), privacy is ensured using DP, directly on the video. This approach allows for visual privacy evaluation, where privacy-utility trade-off is quantified using the $\epsilon$,$\delta$ parameters of DP.} \label{rch}
\end{figure*}

Current literature indicates that privacy preservation, in Video HAR can be achieved either at a model level or directly on the data by modifying its visual content. Model-based approaches usually ensure privacy by leveraging differential privacy (DP) \cite{dwork2006calibrating,dwork2014algorithmic,chaudhuri2011differentially,levy2021learning}. This method provides a theoretical and empirical guarantee of privacy by incorporating noisy mechanisms into the training algorithms, using the privacy parameters $\epsilon$ and $\delta$ \cite{abadi2016deep,luo2024differentially,pichapati2019adaclip,davody2020effect,du2021dynamic}. However, its effectiveness is limited when it comes to post-training privacy analysis such as visual privacy. In the context of video HAR, visual privacy can be define as a model's ability to recognize visual information such as faces, gender, or individuals performing activities. The underlying hypothesis is that diminished performance in these recognition tasks indicates higher visual privacy. As shown in Figure \ref{rch}$(a)$, models trained with DP cannot achieve this level of privacy because the data itself is not directly altered for DP—only the gradient's estimates $g$ are adjusted during training.
\begin{figure*}[tb!]
\includegraphics[width=1\linewidth]{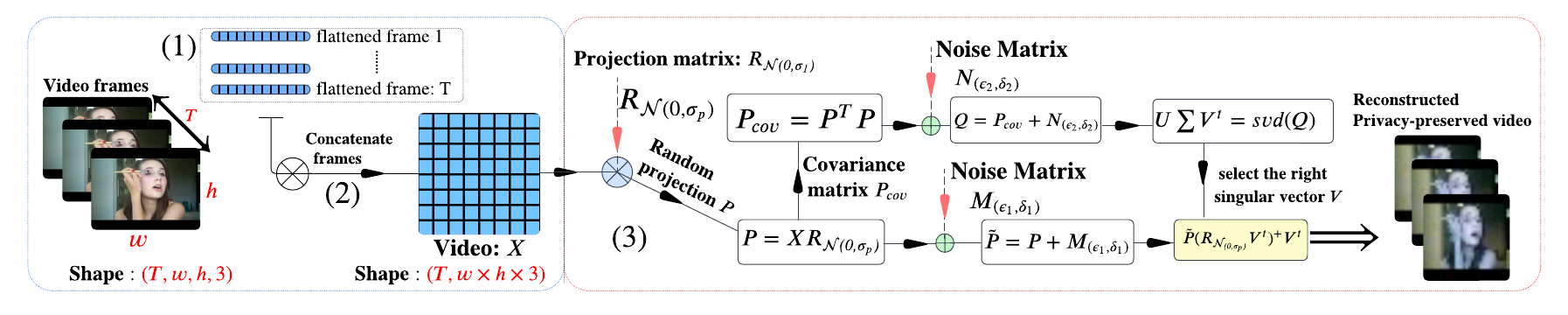}
\caption{Video-DPRP consists of the following components:$(1)$ Each video frame is reshaped and flattened, then concatenated to form a video $X$ of dimension 
$(T,w\times h\times 3)$. $(2)$ A random projection matrix 
$\mathcal{R}_{\mathcal{N}(0,\sigma_{p})}$ reduces $X$ to a lower-dimensional space 
$(T,k)$. $(3)$ Noise is added to both the projected video and its covariance matrix, from which the right singular component 
$V$ of the noisy covariance $Q$ is used to reconstruct a differentially private video (see Section \ref{overview} for details).} \label{project}
\end{figure*}

Conversely, while some data-based approaches utilizing generative adversarial networks (GANs in Figure \ref{rch}$(b)$) offer an affordable means of visual privacy assessment \cite{li2019anonymousnet,ren2018learning,hukkelaas2023deepprivacy2,mirjalili2020privacynet}, the generated videos from these methods may still disclose sensitive visual content \cite{shokri2017membership}, as they are trained on unconstrained real-world data. Additionally GANs, including other video down-sampling and obfuscation approaches \cite{ryoo2017privacy,ren2018learning,ilic2024selective}, lacks the rigorous mathematical privacy guarantees afforded by differential privacy. Theoretical privacy assurance is often overlooked in data-based methods, which typically rely on heuristic approaches, ad-hoc obfuscations, or data transformations. These methods lack transparency in how privacy is preserved and can be vulnerable to reverse-engineering or sophisticated attacks \cite{tekli2023framework,jang2024unveiling}, resulting in mere \textit{security through obscurity}. In contrast, differential privacy is grounded in well-established mathematical principles that provide robust privacy guarantees, irrespective of an adversary's capabilities. Moreover, differential privacy offers clearer privacy explainability in terms of the chances of information leakage, quantified by the $\epsilon$ and $\delta$ parameters \cite{nanayakkara2023chances,bullek2017towards,cummings2021need}.

We identify two key limitations in previous privacy-preserving Video HAR studies: (1) Although DP models provide empirical and theoretical privacy guarantees during model training, their privacy-preserving effect does not extend beyond training. This limitation arises because the data itself remains unaltered, retaining visually sensitive content. Evaluating such data on visual privacy metrics is likely to yield poor results.
(2) While some studies propose data-transformed methods for visual privacy evaluation, these approaches still fail to offer theoretical guarantees of privacy. Recent advancements in differential privacy and random projection present promising solutions. By leveraging a random projection matrix followed by the addition of a noise matrix to the projected data, previous work has demonstrated the feasibility of reconstructing differentially private tabular datasets and images \cite{xu2017dppro,lee2015fast,li2023deep,showkatbakhsh2018privacy,li2024smooth,upadhyay2013random,gondara2020differentially}. However, applying differentially private random projections to a video dataset presents a significant challenge due to the added complexity introduced by the temporal dimension of videos.\\
In this work, we introduce Video-DPRP, a \textbf{Video}-sample-wise \textbf{D}ifferentially \textbf{P}rivate \textbf{R}andom \textbf{P}rojection framework tailored for visual privacy-preserved video reconstruction of HAR datasets. The framework unfolds in several stages: we begin by reshaping each video, as illustrated in Figure \ref{project}. Next, we apply a random projection to the reshaped video using a projection matrix, reducing its dimensionality while preserving its underlying structure. To ensure differential privacy, we add a noise matrix, calibrated with the $(\epsilon, \delta)$ parameters, to both the projected video and its covariance matrix. Finally, by leveraging the right singular vectors from the singular value decomposition (SVD) of the noisy covariance matrix, we reconstruct a video sample, that is both visually and differentially private. \textbf{Ideally, a model trained with videos reconstructed using Video-DPRP is expected to exhibit both high-quality performance in video HAR and strong privacy preservation}. Our contributions are as follows:
\begin{itemize}
    \item We introduce Video-DPRP, a differentially private approach for video reconstruction tailored for video HAR. Video-DPRP provides a theoretical guarantee of differential privacy, while also ensuring visual privacy.
    \item We evaluate the performance of Video-DPRP across both HAR and visual privacy-preserving attributes. For HAR evaluation, we use the UCF101 and HMDB51 datasets. To assess visual privacy-preserving attributes, we utilize the PA-HMDB and the VISPR datasets.
\end{itemize}

%% file: sec/2_formatting.tex
\section{Related Work}\label{sec:rel}
\textbf{Privacy-preserving video HAR.} Privacy in the context of video HAR can be categorized into two main groups: visual privacy and differential privacy.\\
Visual privacy aims to obscure identifiable visual attributes in video content and can be categorized into 3 main groups: obfuscation, adversarial training and downsampling

\textbf{Downsampling}. As an example, Ryo \etal \cite{ryoo2017privacy} proposed an inverse-super-resolution paradigm that learns an optimal set of transformations to generate low-resolution videos from high-resolution inputs. This approach utilizes a down-sampling technique, similar to the methods proposed by \cite{dai2015towards,liu2020indoor}. While this technique is effective, its major drawback lies in the trade-off between achieving accurate activity recognition and maintaining privacy preservation—a trade-off that could be better quantified with a rigorous mathematical bound on privacy. 

\textbf{Obfuscation}. Ren \etal \cite{ren2018learning} presents a data obfuscation method for anonymizing facial images, using a learnable modifier. This approach employs an adversarial training setup, where a generator produces modified versions of facial images, and a discriminator attempts to identify facial features despite the modifications. The end result is a video anonymizer that performs pixel-level modifications to anonymize each person’s face with minimal impact on action detection performance. A similar approach was employed by Zhang \etal \cite{zhang2021multi}, where they first used semantic segmentation to identify the face's region of interest, followed by blurring to achieve face privacy preservation. Additional work on obfuscation has been conducted by Ilic \etal, focusing on appearance-free action recognition using an optical-flow estimator \cite{ilic2022appearance} and selective video obfuscation using random noise \cite{ilic2024selective}. However, obfuscation techniques have a limitation in that they require domain knowledge to effectively identify and obscure the region of interest.

\textbf{Adversarial training}. Beyond video down-sampling and obfuscation, some researchers have developed privacy optimization strategies using adversarial neural networks \cite{wu2020privacy,xiao2020adversarial,pittaluga2019learning,wu2018towards,dave2022spact}. These strategies typically involve a cost function that is minimized for activity recognition, while simultaneously maximized for privacy preservation. A significant drawback of these techniques is their substantial computational resource requirements for reconstructing anonymized videos. In contrast, a more effective approach could be a model-free method capable of reconstructing videos at a considerably lower computational cost.

While visual privacy focuses on \textit{hiding} identifiable visual attributes in sample videos, in differential privacy, a random noise is added to the gradient estimates during a model's training process. This noise is carefully calibrated to ensure that the model can still learn overall patterns and trends, while specific details that could identify a sample video are not leaked. It is important to note that the sample videos themselves are not directly modified; only their gradient estimates are altered during training. Figure \ref{rch}$(a)$ provides a clear illustration of training with differential privacy, specifically detailing a variant of stochastic gradient descent (SGD) known as differential private stochastic gradient descent (DP-SGD) \cite{abadi2016deep}. DP-SGD differs from traditional SGD in that, after computing the per-sample gradient  $g$  it is clipped to a threshold value $C$, resulting in a clipped gradient $g_{clip}$. A Gaussian noise, calibrated with the DP parameters: $\epsilon,\delta$, is then carefully added to the clipped gradient producing the differentially private gradient $g_{dp}$ (details about the DP parameters are provided in Section \ref{overview}). Recently, Luo \etal \cite{luo2024differentially} proposed Multi-Clip DP-SGD, a method designed to achieve video-level differential privacy in HAR. The DP framework is built such that, during model training, shorter video segments, or clips, are sampled from each video, and their gradients are computed and averaged across all the clips of the video. DP-SGD is then applied to the averaged gradient, ensuring differential privacy without additional privacy loss. Although the result is a differential private model, a significant challenge with DP-SGD and other DP learning algorithms is that privacy preservation is confined to the training phase, restricting further visual privacy assessments on the video data beyond training.\\ 
\textbf{Differential Private Random Projection (DPRP).} Previous research introduced DPRP primarily as a \textit{data release} framework, for tabular data \cite{xu2017dppro,gondara2020differentially,kenthapadi2012privacy,ahmed2013random}. For instance Xu \etal \cite{xu2017dppro} employed DPRP for the release of high-dimensional data, while Gondara \etal \cite{gondara2020differentially} adapted DPRP for smaller clinical datasets. In both scenarios, the original dataset is projected into a significantly lower-dimensional space using a random projection matrix, followed by the addition of a noise matrix. This noise matrix is calibrated with the $(\epsilon,\delta)$ parameters to achieve differential privacy. In our approach, we apply DPRP on a per-video-sample basis rather than across the entire dataset, offering more granular privacy control and assessment on activity recognition.

%% file: sec/3_finalcopy.tex
\section{Method Overview}\label{overview}
We begin by introducing key concepts relevant to Video-DPRP, including an initial video transformation mechanism, the theoretical foundations of differential privacy, random projection, and the algorithmic framework of Video-DPRP. This section concludes with preliminary discussions of the privacy guarantees offered by Video-DPRP, which are further detailed in the Appendix.
\subsection{Video Transformation}\label{transform}
A sample video is structured as a $4D$ tensor, ($T$,$w$,$h$,3), consisting of a $1D$ temporal dimension and $2D$ spatial dimensions. The temporal dimension is represented by the number of frames, $T$, in the video sequence, while the spatial dimensions are denoted by the pair $(w, h)$, corresponding to the width and height of each frame. Moreover, each frame contains $3$ color channels (red,green and blue). To facilitate our random projection strategy, we flattened the $2D$ spatial dimensions of each frame from $(1, w, h, 3)$ to $(1, w \times h \times 3)$. Next, we concatenate all the $T$ flattened frames along the temporal axis (the first axis), resulting to $2D$ array $X$ of dimension $(T, w \times h \times 3)$. This concatenation preserves the temporal sequence of the video, with each row of $X$ corresponding to a flattened frame. This step is crucial for our subsequent methodology, and henceforth, we treat each video $X$ as a $2D$ array. 
\subsection{Differential Privacy}
We consider two sample videos, $X$ and $X'$, that differ by a single row, representing neighboring inputs. Intuitively, this means $X$ and $X'$ differ by one frame. Video-DPRP ensures that modifying the pixel values of a single frame does not pose a significant visual privacy risk, nor does it lead to a substantial drop in video HAR performance. This implies that even if an adversary knows the output video, they cannot infer sensitive information about the frame that was modified. We then give a formal definition introduced by Dwork \etal \cite{dwork2006our} and re-calibrated to our context:
\begin{definition}[Differential Privacy]
    A randomized mechanism $\mathcal{M}$, satisfies $(\epsilon,\delta)$-differential privacy if for any two input videos $X$ and $X'$, that differ in only one row (frame), and for all sets of possible outputs $O \in range(\mathcal{M})$, we have:
    \[
\Pr[\mathcal{M}(X) \in O] \leq e^\epsilon \cdot \Pr[\mathcal{M}(X') \in O]+\delta
\]
\end{definition}
\hspace{-0.05\linewidth}In other words, the outcomes of applying the random mechanism $\mathcal{M}$ to the two neighboring videos $X$ and $X'$ differ by at most a factor of $e^{\epsilon}$. The privacy guarantee can fail with a probability of $\delta$. When $\delta=0$, the mechanism operates under pure $\epsilon$-differential privacy.
\subsection{Random Projection}
Random projection is a dimensionality reduction technique that projects data from an initial dimension $d$ to a lower dimension $k$, while preserving pairwise distances between data points (in our case, frames) using a projection matrix $\mathcal{R}$. 
To ensure that the pairwise distances between frames are preserved, the projection matrix must satisfy the Johnson-Lindenstrauss Lemma \cite{johnson1984extensions}.
\begin{lemma}[Johnson-Lindenstrauss \cite{johnson1984extensions}]\label{lemma}
Let $\mathcal{S}$ be a set of $n$ points such that $\mathcal{S} \subset \mathbb{R}^d$, with $\lambda > 0$ and $k = \frac{20 \log n}{\lambda^2}$. There exists a Lipschitz mapping $f: \mathbb{R}^d \to \mathbb{R}^k$ that distorts all pairwise distances by a factor of $1 \pm \lambda$. For any $x, y \in \mathbb{R}^d$, this mapping satisfies the following inequality:
\[
(1 - \lambda) \|x - y\|_2^2 \leq \|f(x) - f(y)\|_2^2 \leq (1 + \lambda) \|x - y\|_2^2
\]
\end{lemma}
\hspace{-0.05\linewidth}Contextually, for a given video $X$, the initial dimension is $d = w \times h \times 3$, where $w$ and $h$ are the width and height of the frames, respectively, and the set of $n$ points corresponds to the number of frames $T$, as discussed earlier in section \ref{transform}. To project the video $X^{T \times d}$, a random projection matrix $\mathcal{R}$ is required, such that the resulting projected video is $P = X\mathcal{R}$. A suitable random projection matrix that satisfies Lemma \ref{lemma} is one whose entries are drawn from a normal distribution with mean $\mu=0$ and variance $\sigma^2=\frac{1}{k}$ (that is, $\mathcal{R}\sim \mathcal{N}(0,\frac{1}{\sqrt{k}})^{d\times k}$).
\subsection{Video-DPRP Algorithm}\label{algo}
The algorithmic framework of Video-DPRP is inspired by the influential work on the Johnson-Lindenstrauss transform \cite{kenthapadi2012privacy}, advancements in image reconstruction \cite{zhang2014image}, and recent developments in the release of small datasets \cite{gondara2020differentially}.
\begin{algorithm}
\caption{Video-DPRP}
\label{alg:example}
\begin{algorithmic}[1]  
\Statex \texttt{\small{\textbf{Input:}}} $D=\{X_1,X_2,....,X_n\}, d \times k, \epsilon,\delta,b$ 
\Statex \texttt{\small\textcolor{black}{/*The dataset D with n videos; the size of the projection matrix $d$$\times$$k$; the privacy parameters $\epsilon$ and $\delta$; the privacy budget allocator $b \in \left]0,1\right[$ */}}
\State \texttt{$\epsilon_1,\delta_1=\epsilon \times b, \delta \times b$}
\State \texttt{$\epsilon_2,\delta_2=\epsilon \times (1-b), \delta \times (1-b)$}
\State \texttt{\small for $X^{T\times d} \in D$ do:}
\State  \texttt{\small~$\mathcal{R}$$\sim$ $\mathcal{N}(0,\frac{1}{\sqrt{k}})^{d\times k}$}\texttt{\small\textcolor{black}{/*projection matrix*/}}
\State \texttt{\small{~$P=X\mathcal{R}$ \textcolor{black}{/*Random projection:\textcolor{blue}{$O(Tdk)$}*/}}}
\State \texttt{\small~$\Tilde{P}=P+M_{(\epsilon_1,\delta_1)}$ \textcolor{black}{/*Noise addition:\textcolor{blue}{$O(Tk)$}*/}}
\State \texttt{\small~$P_{cov}=P^{t}P$\textcolor{black}{/*Covariance matrix:\textcolor{blue}{$O(Tk^2)$}*/}}
\State \texttt{\small~$Q=P_{cov}+N_{(\epsilon_2,\delta_2)}$ \textcolor{black}{/*Noise addition:\textcolor{blue}{$O(k^2)$}*/}}
\State \texttt{\small~$U\sum V^{t}= \texttt{SVD}(Q)$ \textcolor{black}{/*Decomposition:\textcolor{blue}{$O(k^3)$}*/}}
\State \texttt{\small~$ \Tilde{X}=\Tilde{P}\left(\mathcal{R}V^{t}\right)^{+}V^{t}$\textcolor{black}{/*{reconstructed video:}*/}}
\Statex \texttt{\small\textbf{Output:} reshaped video, reshape($\Tilde{X}$)} 
\end{algorithmic}
\end{algorithm}
\textbf{Preliminary:}
 Recall that each video is initially transformed into a $2D$ matrix of dimensions ($T\times d$), where $T$, is the number of frames and $d=w\times h\times 3$ (with $w$ being the width and $h$ the height of a frame).\\
 \underline{\textbf{Privacy parameters:}} All our privacy parameters are derived from a single privacy parameter pair $(\epsilon,\delta)$. To ensure that a given video remains differentially private without significantly compromising its utility, we avoid adding multiple independent noise matrices. Instead, we split the parameters into two sets: one set is used to make the random projection $P$ differentially private $(\epsilon_1,\delta_1)$ and the other set $(\epsilon_2,\delta_2)$ to make the covariance matrix $P_{cov}$, differentially private. Each set is derived using the privacy budget allocator $b \in ]0,1[$ (see \texttt{\small lines 1-2}). The privacy budget is a parameter that controls the total amount of privacy loss allowed, balancing utility with privacy protection. 
 The complete workflow of Video-DPRP is outlined in Algorithm \ref{alg:example}. The time complexity of each step of the algorithm is highlighted in \textcolor{blue}{\texttt{blue}}.
 To begin with, for each video $X^{T \times d}$ in the dataset $D$, we project the video into a lower-dimensional space, using the projection matrix $\mathcal{R}^{d\times k}$ (\texttt{\small lines 4-5}), which satisfies the Johnson-Lindenstrauss Lemma \ref{lemma}. This result to a projected video $P$, of dimension $(T \times k)$. Here, $k$ represents the number of dimensions for the random projection. At this stage, $P$ still contains sensitive information from $X$ and is therefore not differentially private. Differential privacy is ensured by adding a random noise matrix $M_{(\epsilon_{1},\delta_{1})}$ to the projected video (\texttt{\small line 6}), resulting to $\Tilde{P}$. The entries of the random noise matrix are drawn from a Gaussian distribution with mean $\mu=0$ and variance $\sigma^2_{1}$ ($M_{(\epsilon_1,\delta_1)} \sim \mathcal{N}(0,\sigma_1^{2})^{T \times k}$). The variance $\sigma^2_{1}$ is determined using Theorem \ref{theo3}. 
 Differentially private video reconstruction effectively begins at \texttt{\small line 7}, where the covariance matrix $P_{cov}$ of the  projected video $P$, is first computed as a necessary step for reconstruction. The use of the covariance matrix is motivated by principles similar to those in Principal Component Analysis (PCA) \cite{abdi2010principal}, aiming to capture the most significant features of the video within the low-dimensional subspace. 
Similar to \texttt{line 5}, since $P$ is not differentially private, its covariance matrix $P_{cov}$ is also not. To achieve differential privacy, a random noise matrix $N_{(\epsilon_2,\delta_2)}$ is added to $P_{cov}$, resulting in a noisy covariance matrix $Q$ (\texttt{\small line 8}) of dimension ($k \times k$). In the same way, the entries of $N_{(\epsilon_2,\delta_2)}$ are drawn from a Gaussian distribution with mean $\mu=0$ and variance $\sigma_2^{2}$ ($N{(\epsilon_2,\delta_2)} \sim \mathcal{N}(0,\sigma_2^{2})^{k \times k}$). The variance $\sigma_{2}^2$ is determined using Theorem \ref{theo4}. To proceed, the noisy covariance matrix $Q$ is subjected to a singular value decomposition (SVD), which decomposes $Q$ into three matrices: $U\Sigma V^{t}$ (\texttt{\small line 9}), where $U$ and $V^{t}$ (denoting the transpose of $V$) are orthogonal matrices each of dimensions ($k \times k$), and $\Sigma$ is a diagonal matrix containing the singular values. Following the approach of \cite{gondara2020differentially}, we only use the right singular component $V^{t}$, the random projection matrix $\mathcal{R}$, and the differentially private projected video $\Tilde{P}$ for video reconstruction of $\Tilde{X}$ (\texttt{line 10}). We use the Moore-Penrose pseudoinverse (denoted by \texttt{+}) of $\mathcal{R}V^{t}$ because $\mathcal{R}V^{t}$ is not a squared matrix and may not be invertible. $\Tilde{X}$ has dimensions $(T\times d)$ and is ultimately reshaped back to its original video format $(T,w,h,3)$.\\
\underline{\textbf{Time complexity:}} Given that the algorithm processes $n$ videos independently, the overall time complexity for the entire dataset $D$ is $O(n(Tdk+Tk^2+k^3))$. This complexity shows that the algorithm scales linearly with the number of videos $n$, and is influenced by both the number of frames 
$T$ and the dimensionality $d$. The cubic term $k^3$ becomes dominant when the projection dimension $k$ is large.
\subsection{Privacy Guarantee of Video-DPRP}\label{guarantee}
Differential privacy is applied at two stages in Algorithm \ref{alg:example}: (i) to ensure that the projected video $P$ is differentially private, and (ii) to make the covariance matrix $P_{cov}$ differentially private. To establish the privacy guarantee of Video-DPRP, we must demonstrate that both stages meet differential privacy requirements. The proofs rely on two supporting theorems from \cite{tudifferentially, gondara2020differentially}, which are included here for completeness, with details provided in the appendix.
\begin{theorem}[Privacy of projected video $P$]\label{theo3}
    Let $\epsilon_1>0$ and $0<\delta_1<\frac{1}{2}$. Consider a randomized Gaussian projection matrix $\mathcal{R\sim N}(0,{1}/{\sqrt{k}})^{d \times k}$. Then, the noisy projection $\Tilde{P}=X\mathcal{R}+M_{(\epsilon_1,\delta_1)}$, where $M_{(\epsilon_1,\delta_1)}$ is a $(T\times k) $ Gaussian matrix with entries drawn from $\mathcal{N}(0,\sigma_1^2)$, is $(\epsilon_1,\delta_1)$-differentially private, with:
   \footnotesize \[
    \sigma_1= \theta\sigma_p\sqrt{k+2\sqrt{klog({2}/{\delta_1})}+2log({2}/{\delta_1})}{\sqrt{2(log(1/2\delta_1)+\epsilon_1)}}/{\epsilon_1}
    \]    
\end{theorem}
\hspace{-0.05\linewidth}Where $\sigma_p={1}/{\sqrt{k}}$, and $\theta$ denotes the $L_{2}$ sensitivity bound of the input. The variables are consistent with those defined in Section \ref{algo} to maintain uniformity.\\
\hspace{-0.05\linewidth}\underline{\textbf{The $L_{2}$ sensitivity $\theta$:}} For the input $f(X)=X\mathcal{R}$ where $X$ represents the video with pixel values ranging from $[0,255]$ and 
$\mathcal{R}$ is a random matrix, the $L_{2}$ sensitivity $\theta$ is proportional to the maximum change in $X$, scaled by the norm of 
$\mathcal{R}$. This norm typically takes the value $1/\sqrt{k}$. Since the $L_{2}$ sensitivity of $X$ is $|255-0|$, we define $\theta$ as $\theta=255/\sqrt{k}$. Where $\left|.\right|$ denotes the absolute value. More details are provided in the appendix section.
\begin{theorem}[Privacy of covariance matrix $P_{cov}$]\label{theo4}
  The mechanism defined by $Q$=$P_{cov}$+$N_{(\epsilon_2,\delta_2)}$, where $N_{(\epsilon_2,\delta_2)}$ is a Gaussian matrix with entries drawn from $\mathcal{N}(0,\sigma_2)$, is $(\epsilon_2,\delta_2)$-differentially private, provided that $\epsilon_2>0$ and $\delta_2<1/2$. Where $\sigma_2$ $=$$\theta\sqrt{\frac{\sqrt{2log(1.25)/\delta_2}}{\epsilon_2}}$.
\end{theorem}
\hspace{-0.05\linewidth}By applying the principle of sequential composition \cite{dwork2014algorithmic}, each video $X$ is $(\epsilon,\delta)$-differentially private as a result of the combination of two differentially private mechanisms in  Algorithm \ref{alg:example}. Where $\epsilon=\epsilon_1+\epsilon_2$ and $\delta=\delta_1+\delta_2$.
\setlength{\fboxsep}{0.7pt}
\begin{table*}[ht]
\centering
\footnotesize
\begin{tabular}{lcccc|cccc}
\toprule
\multirow{4}{*}{\textbf{Method}} 
& \multicolumn{2}{c}{\textbf{Raw Test set} Top-1($\uparrow$)} & \multicolumn{2}{c}{\textbf{Reconstructed Test set} Top-1($\uparrow$)} & \multicolumn{2}{c}{\textbf{Raw Test set PA-HMDB}}&\multicolumn{2}{c}{\textbf{Raw Test set VISPR}}\\
\cmidrule(lr){2-3} \cmidrule(lr){4-5} \cmidrule(lr){6-7}\cmidrule(lr){8-9} 
&\text{UCF101} & \multicolumn{1}{c}{\text{HMDB51}}
& \text{UCF101} & \multicolumn{1}{c}{\text{HMDB51}}
& \text{Top-1 {($\uparrow$)}} & \text{cMAP {($\downarrow$)}} & \text{cMAP {($\downarrow$)}}& \text{F1 {($\downarrow$)}} \\
\midrule
ISR$_{(32\times 24)}$\cite{ryoo2017privacy}& 49.65$\scriptstyle{\pm0.22}$ & 35.66$\scriptstyle{\pm0.10}$ &45.14$\scriptstyle{\pm0.53}$ & 28.97$\scriptstyle{\pm0.09}$ & 38.71$\scriptstyle{\pm1.22}$&58.26$\scriptstyle{\pm0.13}$& 53.60$\scriptstyle{\pm0.87}$ & 49.14$\scriptstyle{\pm0.09}$  \\
ISR$_{(16\times 12)}$\cite{ryoo2017privacy}& 18.34$\scriptstyle{\pm0.02}$ & 19.47$\scriptstyle{\pm0.04}$ &24.94$\scriptstyle{\pm0.20}$ & 12.64$\scriptstyle{\pm0.01}$ &25.11$\scriptstyle{\pm0.62}$&40.01$\scriptstyle{\pm0.17}$& 43.27$\scriptstyle{\pm0.25}$ & 45.00$\scriptstyle{\pm0.73}$  \\
V-SAM\cite{hukkelaas2023realistic}&  17.32$\scriptstyle{\pm0.30}$ & 14.72$\scriptstyle{\pm0.12}$&10.02$\scriptstyle{\pm1.48}$& 12.03$\scriptstyle{\pm0.91}$ & 15.31$\scriptstyle{\pm0.54}$& 40.39$\scriptstyle{\pm0.38}$&44.64$\scriptstyle{\pm0.09}$& \color{red}{\textbf{39.97$\scriptstyle{\pm0.16}$}} \\
Face Anonymizer\cite{ren2018learning}&  32.05$\scriptstyle{\pm0.49}$ & 19.04$\scriptstyle{\pm0.24}$ &21.62$\scriptstyle{\pm0.35}$& 21.13$\scriptstyle{\pm0.69}$ &17.04$\scriptstyle{\pm0.03}$&41.18$\scriptstyle{\pm1.09}$& 44.00$\scriptstyle{\pm0.63}$& 51.43$\scriptstyle{\pm0.39}$ \\
SPAct\cite{dave2022spact}& \underline{60.82$\scriptstyle{\pm0.33}$} & \color{red}{\textbf{41.29$\scriptstyle{\pm0.01}$}} & -& - &44.13$\scriptstyle{\pm0.73}$ &60.55$\scriptstyle{\pm0.75}$ &56.71$\scriptstyle{\pm0.18}$ & 47.61$\scriptstyle{\pm0.11}$\\
ALF\cite{wu2020privacy}& 56.27$\scriptstyle{\pm0.91}$ & 32.04$\scriptstyle{\pm0.56}$ &- & - &43.73$\scriptstyle{\pm0.82}$ &40.29$\scriptstyle{\pm0.03}$&55.09$\scriptstyle{\pm1.49}$&43.08$\scriptstyle{\pm1.02}$\\
Deepprivacy\cite{hukkelaas2023deepprivacy2}& 16.72$\scriptstyle{\pm0.36}$ & 11.54$\scriptstyle{\pm0.05}$ & 14.95$\scriptstyle{\pm0.58}$&11.69$\scriptstyle{\pm0.90}$ &18.77$\scriptstyle{\pm0.13}$ &\underline{39.76$\scriptstyle{\pm0.83}$}&\underline{42.06$\scriptstyle{\pm0.28}$} & 41.27$\scriptstyle{\pm0.50}$\\
Appearance free\cite{ilic2022appearance}& 30.02$\scriptstyle{\pm0.07}$ & 15.67$\scriptstyle{\pm0.14}$ &14.22$\scriptstyle{\pm0.16}$&10.29$\scriptstyle{\pm0.06}$ &19.60$\scriptstyle{\pm0.02}$&-& - & -\\
Selective privacy\cite{ilic2024selective}& {58.97$\scriptstyle{\pm0.11}$} &38.27$\scriptstyle{\pm0.01}$&45.10$\scriptstyle{\pm0.15}$&\underline{30.09$\scriptstyle{\pm0.06}$} & 42.56$\scriptstyle{\pm0.54}$&-&-&- \\
Face blurring\cite{jaichuen2023blur}& 51.07$\scriptstyle{\pm0.63}$ & 37.98$\scriptstyle{\pm0.21}$ &40.01$\scriptstyle{\pm0.64}$&28.81$\scriptstyle{\pm0.01}$&37.00$\scriptstyle{\pm0.44}$&42.13$\scriptstyle{\pm0.68}$& 47.04$\scriptstyle{\pm0.33}$ & 52.34$\scriptstyle{\pm0.27}$\\
\rowcolor{black!15}
Raw data (no privacy) & {85.77$\scriptstyle{\pm0.18}$} &{59.24$\scriptstyle{\pm0.65}$}&{85.77$\scriptstyle{\pm0.18}$}& {59.24$\scriptstyle{\pm0.65}$} &{65.05$\scriptstyle{\pm1.17}$} & {70.13$\scriptstyle{\pm0.59}$}&{64.18$\scriptstyle{\pm0.06}$} & {69.10$\scriptstyle{\pm0.59}$}\\
\toprule
\rowcolor{gray!5}
\textbf{Video-DPRP}$_{(\epsilon=2,\delta=10^{-4})}$& 55.16$\scriptstyle{\pm0.59}$ &36.49$\scriptstyle{\pm0.79}$&38.76$\scriptstyle{\pm0.11}$& 27.95$\scriptstyle{\pm0.01}$ &39.25$\scriptstyle{\pm0.15}$&\color{red}{\textbf{39.42$\scriptstyle{\pm0.62}$}} & \color{red}{\textbf{41.89$\scriptstyle{\pm0.02}$}} &\underline{40.03$\scriptstyle{\pm0.28}$}\\
\rowcolor{gray!5}
\textbf{Video-DPRP}$_{(\epsilon=5,\delta=10^{-4})}$& 58.58$\scriptstyle{\pm0.16}$ & 38.37$\scriptstyle{\pm0.09}$&\underline{45.20$\scriptstyle{\pm0.02}$}  & 29.06$\scriptstyle{\pm0.77}$  &\underline{44.86$\scriptstyle{\pm0.02}$}&48.75$\scriptstyle{\pm0.07}$& 50.11$\scriptstyle{\pm0.63}$ & 53.77$\scriptstyle{\pm0.30}$\\
\rowcolor{gray!5}
\textbf{Video-DPRP}$_{(\epsilon=8,\delta=10^{-4})}$& \color{red}{\textbf{61.69$\scriptstyle{\pm0.07}$}} & \underline{40.00$\scriptstyle{\pm0.71}$}&\color{red}{\textbf{50.00$\scriptstyle{\pm0.28}$}}& \color{red}{\textbf{32.13$\scriptstyle{\pm0.31}$}} & \color{red}{\textbf{51.07$\scriptstyle{\pm0.73}$}}& 53.00$\scriptstyle{\pm0.01}$& 56.10$\scriptstyle{\pm0.20}$ & 55.12$\scriptstyle{\pm0.03}$\\
\bottomrule
\end{tabular}
\caption{Comparison with different visual privacy techniques, including data-obfuscation, adversarial training and video anonymization using GANS. cMAP and F1 metrics are for \textit{privacy evaluation} while Top-1 is for \textit{action evaluation}. Results are reported on UCF101 \cite{soomro2012ucf101}, HMDB51 \cite{kuehne2011hmdb},PA-HMDB \cite{wu2020privacy} and VISPR \cite{orekondy2017towards}. The best results are in \textcolor{red}{\textbf{red}}, while the second best are \underline{underlined}.}\label{visual}
\end{table*}
\begin{table*}[ht]
\hspace{-0.02\linewidth}
\centering
\footnotesize
\begin{tabular}{lcccccc|ccc}
\toprule
\multirow{3}{*}{\textbf{Method}} 
& \multicolumn{3}{c}{\textbf{Raw Test set UCF101} (\text{Top-1 {($\uparrow$)}})} & \multicolumn{3}{c}{\textbf{Raw Test set HMDB51} (\text{Top-1 {($\uparrow$)}})} & \multicolumn{3}{c}{\textbf{Raw Test set PA-HMDB} (\text{Top-1 {($\uparrow$)}})}\\
\cmidrule(lr){2-4} \cmidrule(lr){5-7} \cmidrule(lr){8-10}
&\textbf{$\epsilon=2$} & \multicolumn{1}{c}{\textbf{$\epsilon=5$}} & \textbf{$\epsilon=8$} 
& \textbf{$\epsilon=2$} & \multicolumn{1}{c}{\textbf{$\epsilon=5$}} & \textbf{$\epsilon=8$}
& \textbf{$\epsilon=2$} & \textbf{$\epsilon=5$} & \textbf{$\epsilon=8$}
 \\
\midrule
\hspace{-0.01\linewidth}DP-SGD\cite{abadi2016deep} &$\scriptstyle{25.54\pm0.33}$ & $\scriptstyle{37.24\pm0.26}$ &$\scriptstyle{45.32\pm0.86}$ & $\scriptstyle{14.18\pm0.06}$ & $\scriptstyle{30.09\pm0.18}$&$\scriptstyle{32.16\pm1.85}$&$\scriptstyle{15.34\pm2.15}$ & $\scriptstyle{25.70\pm1.97}$ & $\scriptstyle{29.90\pm0.04}$\\
\hspace{-0.01\linewidth}MultiClip-DP$_{\text{(3 clips)}}$\cite{luo2024differentially} & $\scriptstyle{44.07\pm0.18}$ & \underline{$\scriptstyle{70.03\pm0.13}$} &\underline{$\scriptstyle{72.03\pm0.71}$} & {$\scriptstyle{36.11\pm0.25}$} &{$\scriptstyle{48.00\pm0.05}$}&\underline{$\scriptstyle{50.98\pm0.66}$}&$\scriptstyle{37.06\pm0.24}$&\underline{$\scriptstyle{45.16\pm0.05}$}& \underline{$\scriptstyle{52.73\pm0.46}$}\\
\rowcolor{black!15}
\rowcolor{gray!5}
\hspace{-0.01\linewidth}\textbf{Video-DPRP} &\underline{$\scriptstyle{55.16\pm0.59}$} &{$\scriptstyle{58.58\pm0.16}$}&{$\scriptstyle{61.69\pm0.07}$} &\underline{$\scriptstyle{36.49\pm0.79}$} &\underline{$\scriptstyle{38.37\pm0.09}$}&{$\scriptstyle{40.00\pm0.71}$}&\underline{$\scriptstyle{39.25\pm0.15}$} &$\scriptstyle{44.86\pm0.02}$ & $\scriptstyle{51.07\pm0.73}$\\
\rowcolor{gray!5}
\hspace{-0.01\linewidth}\textbf{Video-DPRP}$_{\text{(3 clips)}}$ & \color{red}{$\scriptstyle{60.11\pm0.10}$} & \color{red}{$\scriptstyle{70.87\pm0.03}$}& \color{red}{$\scriptstyle{74.06\pm0.38}$}& \color{red}{$\scriptstyle{42.72\pm0.44}$} &\color{red}{$\scriptstyle{49.63\pm0.95}$} & \color{red}{$\scriptstyle{51.63\pm0.53}$}& \color{red}{$\scriptstyle{41.08\pm0.04}$} & \color{red}{$\scriptstyle{48.17\pm0.42}$} & \color{red}{$\scriptstyle{54.98\pm0.86}$}\\
\bottomrule
\end{tabular}
\caption{Comparison with differential private training methods and Video-DPRP on action recognition, for $\epsilon$ $\in$ $\{2, 5, 8\}$ and $\delta$$=$$10^{-4}$. The best results are in \textcolor{red}{red} , and the second best are \underline{underlined}. }\label{diffpriv}
\end{table*}

%% file: sec/4_experiment.tex
\section{Experiments}
\subsection{Datasets}
We adopt \textbf{PA-HMDB} \cite{wu2020privacy} and \textbf{VISPR} \cite{orekondy2017towards} for visual privacy assessment, and \textbf{UCF101} \cite{soomro2012ucf101} and \textbf{HMDB51} \cite{kuehne2011hmdb} for HAR, as these are commonly used datasets in the literature.\\ 
\textbf{PA-HMDB}\cite{wu2020privacy} is a dataset containing 515 videos with video-level action annotations and frame-wise visual privacy annotations, including privacy attributes such as \textit{skin color, face, gender, nudity}, and \textit{relationship}. The dataset covers 51 action classes.\\
\textbf{VISPR} \cite{orekondy2017towards} is an image dataset designed for visual privacy research. It contains various personal attributes similar to those in HMDB51. The dataset comprises $10,000$ training images, $4,100$ validation images, and $8,000$ test images.\\
\textbf{UCF101} \cite{soomro2012ucf101} and \textbf{HMDB51} \cite{kuehne2011hmdb} are both HAR datasets, containing $101$ and $51$ action classes, respectively. For both datasets, all results are reported on split-1, which includes $9,537$ training videos and $3,783$ test videos for UCF101, and $3,570$ training videos and $1,530$ test videos for HMDB51. Further details of the datasets are provided in the Appendix.
\subsection{Implementation details}\label{details}
Many deep learning models incorporate Batch Normalization (Batch Norm) layers. However, such models are not compatible with differentially private training methods like DP-SGD \cite{abadi2016deep} or MultiClip-DP-SGD \cite{luo2024differentially} (abbreviated to MultiClip-DP in Table \ref{diffpriv}), as Batch Norm requires calculating the mean and standard deviation for each mini-batch, introducing dependencies between samples and violating the principles of differential privacy. For fair comparison across all our results in video HAR, we require a model with a different type of normalization layer. Therefore, we use the PyTorch implementation of the Multiscale Vision Transformer (MViT-B$_{(16\times 4)}$) \cite{fan2021multiscale}, which employs Layer Normalization \cite{ba2016layer} and is pre-trained on the large-scale Kinetics-400 dataset \cite{carreira2017quo}. For each video, we randomly crop a clip consisting of 16 frames, with each frame resized to a shape of $(224, 224, 3)$. In the case of Video-DPRP$_{\text{(3 clips)}}$ and MultiClip-DP$_{\text{(3 clips)}}$ (see Table \ref{diffpriv}), we crop $3$ clips and apply the same pre-processing as described above. The optimization is performed using stochastic gradient descent (SGD) \cite{bottou2010large} with a learning rate of $lr = 0.01$, a batch size of 8, and 50 training epochs.\\
\underline{\textbf{Set-up of Video-DPRP:}} We use video samples reconstructed by Video-DPRP as inputs for our training. In line with Algorithm \ref{alg:example}, we set the dimensions of the projection matrix to $d \times k$, where $d = 320 \times 240 \times 3$ and $k = 32 \times 32 \times 3$. Note that $d$ corresponds to the dimensions of a frame from the original video (as described in Section \ref{algo}) and is therefore fixed to the value defined above by default. We set the privacy budget allocator $b$ to $0.8$, meaning that $80\%$ of the privacy budget is allocated to making the random projection $P$, differentially private (see \texttt{\small line 6}) while the remaining $20\%$ (i.e, $1-b$) is used to ensure the covariance matrix $P_{cov}$ is differentially private (see  \texttt{\small line 8} of Algorithm \ref{alg:example}).

For the differentially private training of DP-SGD \cite{abadi2016deep} and MultiClip-DP$_{\text{(3 clips)}}$ \cite{luo2024differentially}, we use the PyTorch Opacus library \cite{yousefpour2021opacus}, which includes a privacy budget accountant to track the differentially private parameters $(\epsilon, \delta)$ during training. For fair comparison and simplicity across all differentially private techniques (i.e., DP-SGD \cite{abadi2016deep}, MultiClip-DP$_{\text{(3 clips)}}$ \cite{luo2024differentially} and Video-DPRP), we set the privacy parameter $\delta$$=$$10^{-4}$ and only vary $\epsilon$. All experiments were conducted on an NVIDIA RTX A6000 GPU.  For a comparative analysis with state-of-the-art (SOTA) visual privacy techniques, such as \textbf{Obfuscation} and \textbf{Anonymization}, we replicate their techniques following the authors' descriptions. 
Further details about SOTA techniques are provided below.\\
\textbf{Obfuscation Method:}\\
\underline{Inverse Super Resolution (ISR)} \cite{ryoo2017privacy}: For the initial set of raw videos, we begin by performing average downsampling from a resolution of $320 \times 240$ to resolutions of $32 \times 24$ and $16 \times 12$. We then apply a series of random transformations to the downsampled videos, as described in \cite{ryoo2017privacy}, from the set S$=\{\text{shifting},\text{scaling},\text{rotation}\}$. The transformed videos are then used as training input for our model.\\
\underline{Face Blurring} \cite{jaichuen2023blur}: We employ a pretrained YOLOv3 model \cite{redmon2016you} for face detection on the original videos, leveraging its initial training on the Wider Face dataset \cite{yang2016wider}. After detecting the faces, we apply Gaussian blurring to the detected bounding boxes using a kernel size of $k=21$ and a standard deviation of $\sigma=10$ to ensure consistency with previous works \cite{wu2020privacy, ilic2024selective}.\\
\underline{Appearance free Privacy} \cite{ilic2022appearance}:
It utilizes a state-of-the-art optical flow estimator \cite{teed2020raft} to compute optical flow between pairs of subsequent frames in a video. A noise frame is then warp using the set of optical flows to obtain new frames, that are devoid of appearance clues while retaining motion flow like the original video. We use the appearance-free UCF101 dataset (also known as AFD101 \cite{ilic2022appearance}) provided by the authors and synthesize the other video dataset according to the authors' original code.\\
\textbf{Anonymization method:} Given the frames of an original video, we generate an anonymized video by applying pre-trained adversarial networks from three previous works: (i) full-body anonymization with \underline{DeepPrivacy} \cite{hukkelaas2023deepprivacy2}, (ii) a video face anonymizer (\underline{Face Anonymizer} \cite{ren2018learning}), and (iii) a  Variational Surface-Adaptive Modulator (\underline{V-SAM} \cite{hukkelaas2023realistic}).\\
\textbf{Adversarial training method:} We also compare our results with two adversarial training frameworks: one that utilizes a privacy budget (\underline{ALF} \cite{wu2020privacy}) and another that leverages self-supervised learning (\underline{SPAct}\cite{dave2022spact}). For the latter, we use MViT-B$_{(16\times 4)}$ as our target action classification model (referred to as $f_{T}$ in the original paper \cite{dave2022spact}), while keeping the rest of the set-up consistent with the original paper.\\
\textbf{Differentially private training methods:} For the differentially private training of \underline{DP-SGD}\cite{abadi2016deep} and \underline{MultiClip-DP}$_{(3~\text{clips})}$\cite{zhang2021multi}, we use a clipping norm of $C$=$0.4$. Details of these training strategies can be found in Section \ref{sec:rel}. The \textit{3 clips} in MultiClip-DP$_{(3~\text{clips})}$ refers to averaging the gradient across 3 clips of a given video before adding random noise, as described in \cite{zhang2021multi}.
\begin{table*}[ht]
\footnotesize
  \centering
  \begin{minipage}{0.32\textwidth}
    \centering
    \begin{tabular}{lcc}
      \toprule
      \multirow{4}{*}{}
      &\multicolumn{2}{c}{PA-HMDB}\\
      \midrule
      \underline{$\textbf{Dimension}~k$} & \textbf{Action} $~\uparrow$ &\textbf{Privacy}$~\downarrow$ \\
      
      $20$ $\times$ $20$ $\times$ $3$ & 24.60$\scriptstyle{\pm1.67}$ &\textbf{32.46$\scriptstyle{\pm0.75}$}  \\
      $24$ $\times$ $32$ $\times$ $3$ & 28.03$\scriptstyle{\pm0.07}$ & \underline{40.17$\scriptstyle{\pm1.22}$} \\
      $32$ $\times$ $32$ $\times$ $3$ & 51.07$\scriptstyle{\pm0.73}$ & 53.00$\scriptstyle{\pm0.01}$ \\
      
      $50$ $\times$ $50$ $\times$ $3$ & \underline{67.18$\scriptstyle{\pm0.49}$} & 56.96$\scriptstyle{\pm0.07}$ \\
      
      $64$ $\times$ $80$ $\times$ $3$ & \textbf{70.10$\scriptstyle{\pm0.14}$} & 60.18$\scriptstyle{\pm0.33}$ \\
      \bottomrule
    \end{tabular}
    \caption{Action (Top-1) and privacy (cMAP) scores on \textbf{PA-HMDB}\cite{wu2020privacy} for different lower dimensions $k$. The best result is highlighted in \textbf{bold}, and the second best is \underline{underlined}.}\label{dimensionality}
  \end{minipage}
  \hfill
  \begin{minipage}{0.32\textwidth}
    \centering
    \begin{tabular}{lcc}
      \toprule
      \multirow{4}{*}{}
      &\multicolumn{2}{c}{PA-HMDB}\\
      \midrule
      \underline{$\textbf{Budget}~b$} & \textbf{Action}$~\uparrow$ &\textbf{Privacy}$~\downarrow$ \\
      
      $~~~0.2$ & 37.80$\scriptstyle{\pm0.92}$ & \textbf{29.02$\scriptstyle{\pm0.12}$} \\
      
      $~~~0.4$ & 40.26$\scriptstyle{\pm0.17}$ & \underline{36.73$\scriptstyle{\pm1.01}$} \\
      
      $~~~0.5$ & 43.80$\scriptstyle{\pm0.85}$ & 44.27$\scriptstyle{\pm0.49}$ \\
      $~~~0.6$ & \underline{46.39$\scriptstyle{\pm0.22}$} & 50.01$\scriptstyle{\pm0.14}$ \\
      $~~~0.8$ & \textbf{51.07$\scriptstyle{\pm0.73}$} & 53.00$\scriptstyle{\pm0.01}$ \\
      \bottomrule
    \end{tabular}
    \caption{Action (Top-1) and privacy (cMAP) scores on \textbf{PA-HMDB}\cite{wu2020privacy} for different privacy budget $b$. The best result is highlighted in \textbf{bold}, and the second best is \underline{underlined}.}\label{budget}
  \end{minipage}
  \hfill
  \begin{minipage}{0.32\textwidth}
    \centering
    \begin{tabular}{lcc}
      \toprule
      \multirow{4}{*}{}
      &\multicolumn{2}{c}{Reconstruction (sec/Video)}\\
      \midrule
      \underline{$\textbf{Methods}$} & \textbf{UCF101} &\textbf{HMBD51} \\
      
      V-SAM\cite{hukkelaas2023deepprivacy2} & 33.12 & 35.07 \\
      
      ISR$_{(32\times 24)}$\cite{ryoo2017privacy}&\textbf{19.24} & \textbf{18.97}\\
      
      Appearance free \cite{ilic2022appearance} & 23.74& 21.60 \\
      Face blurring \cite{jaichuen2023blur} & 26.08 & 24.49 \\
      Video-DPRP(ours) & \underline{20.32} & \underline{19.84} \\
      \bottomrule
    \end{tabular}
    \caption{Reconstruction time per video (in seconds) for \textbf{UCF101} \cite{soomro2012ucf101} and \textbf{HMDB51} \cite{kuehne2011hmdb}. The best (lowest) time is highlighted in \textbf{bold}, and the second best is \underline{underlined}.}\label{reconst}
  \end{minipage}
\end{table*}
\subsection{Evaluation Metrics and Protocols}\label{protocols}
\textbf{Metrics:} Action recognition evaluation is conducted using the Top-1 accuracy metric, following prior work \cite{hara2018can, luo2024differentially, hara2017learning}. For visual privacy recognition, considered as a multi-label image classification task due to the presence of multiple privacy attributes per image, we use the class-wise mean average precision (cMAP) \cite{orekondy2017towards} and the class-wise F$1$-score. All results are reported as percentages, averaged over three runs, with both their mean and variance provided. In our tables, $\uparrow$ denotes metrics where higher values are better, while $\downarrow$ indicates that lower values are better.\\
\textbf{Protocols:} Apart from \textbf{Adversarial training} methods, which ensure privacy directly during training, we apply two evaluation protocols for video HAR with visual privacy techniques.   \underline{Protocol 1} evaluates on the raw test set X$_{raw}^{test}$ of dataset X$\in$ $\{\textbf{UCF101}, \textbf{HMDB51}, \textbf{PA-HMDB}\}$, after training our model on the corresponding reconstructed train set X$_{reconst}^{train}$ using a method \textit{reconst} $\in$ $\{\textbf{Obfuscation}, \textbf{Anonymization}, \textbf{Video-DPRP}\}$. Here, \textbf{Obfuscation} and \textbf{Anonymization} refer to all obfuscation and anonymization techniques described in Section \ref{details}. It is important to note that for evaluation on the \textbf{PA-HMDB} dataset, we use \textbf{HMDB51}$\setminus$\{\textbf{PA-HMDB}\} as our training set. This means that all video samples present in \textbf{HMDB51} but not in \textbf{PA-HMDB} are used for training in this scenario. \underline{Protocol 2} evaluates on the reconstructed test set X$_{reconst}^{test}$ of dataset X, after training on X$_{reconst}^{train}$, using method \textit{reconst}. Accordingly, no results are provided for adversarial training methods in the \textit{reconstructed test set} column of Table \ref{visual}. \textbf{Protocol 1} assesses the model's robustness in real-world scenarios where obfuscation or anonymization might not be applied, while testing on the reconstructed data (\textbf{Protocol 2}) measures performance consistency under privacy-preserving transformations, validating model adaptability across both standard and privacy-focused settings.  

In Table \ref{diffpriv}, we restrict the analysis of video HAR to differentially private training methods: DP-SGD \cite{abadi2016deep} and MultiClip-DP$_{\text{(3 clips)}}$ \cite{luo2024differentially}, alongside Video-DPRP, as these are the only methods that incorporate differential privacy.\\
For visual privacy evaluation, we begin by training our model on the training set of \textbf{VISPR}, formulating the task as a multi-label image classification problem due to the multiple privacy attributes per image. We then use the annotated video frames from \textbf{PA-HMDB} as our test set. This is considered a cross-dataset evaluation protocol, as outlined in \cite{dave2022spact}. We also evaluate on the test set of \textbf{VISPR}, as reported in Table \ref{visual}. 
We can observe from Table \ref{visual} that \textbf{Video-DPRP} provides competitive results, highlighted in {\textbf{bold}}, when compared to SOTA privacy-preserving techniques in both activity recognition and visual privacy preservation. Notably, the performance of \textbf{Video-DPRP} with $\epsilon$$=$$8$ and $\delta$$=$$10^{-4}$ (i.e Video-DPRP$_{(\epsilon=8,\delta=10^{-4})}$, in Table \ref{visual}) shows a significant improvement. However, Video-DPRP$_{(\epsilon=8,\delta=10^{-4})}$ shows a slight performance drop of $\textbf{1.29\%}$ in activity recognition on the HMDB51 dataset compared to SPAct\cite{dave2022spact}, which achieved a baseline accuracy of \textbf{41.29\%}. In terms of visual privacy, we observe that Video-DPRP achieved a cMAP score of \textbf{39.76\%} on PA-HMDB51 and \textbf{42.06\%} on VISPR (with $\epsilon$$=$$2$), outperforming state-of-the-art methods such as ISR$_{(32\times24)}$\cite{ryoo2017privacy} and V-SAM\cite{hukkelaas2023realistic}. Despite yielding decent scores on visual privacy, anonymization methods such as DeepPrivacy\cite{hukkelaas2023deepprivacy2}, V-SAM\cite{hukkelaas2023realistic} and Face Anonymizer \cite{ren2018learning} as well as obfuscation method like ISR$_{(16\times 12)}$\cite{ryoo2017privacy}, struggle to achieve good utility performance on HAR, with results dropping as low as $\textbf{12.00\%}$. Intuitively, DeepPrivacy\cite{hukkelaas2023deepprivacy2}, V-SAM\cite{hukkelaas2023realistic} and Face Anonymizer\cite{ren2018learning} generate a \textit{modified} version of the original video, which often fails to consistently preserve the motions of individuals involved in the activity. We conclude that while the above anonymization methods yield good privacy results, they may not be suitable for utility analysis in Video HAR. For obfuscation techniques, we argue that the visual content may be so \textit{obscured} that models struggle to effectively identify activities. In contrast, Video-DPRP strikes a balance between utility and privacy, even for varying values of $\epsilon$ $\in$ $\{2,5,8\}$. We do not report the privacy results for Appearance-Free \cite{ilic2022appearance} and Selective Privacy \cite{ilic2024selective}, as both methods rely on optical flow between successive frames in videos for obfuscation, which is not applicable in our experiment since we use VISPR as the primary training set for privacy evaluation.  In Table \ref{diffpriv}, we use DP-SGD \cite{abadi2016deep} as a baseline method and compare our results with MultiClip-DP$_{(\text{3 clips})}$ \cite{luo2024differentially}. It is important to note that the results for MultiClip-DP$_{(\text{3 clips})}$ \cite{luo2024differentially} are based on our own experiments, as the original code was not available. With 3 $clips$ per sample video, Video-DPRP$_{(\text{3 clips})}$ provides competitive results when compared to MultiClip-DP$_{(\text{3 clips})}$, achieving Top-1 accuracy of \textbf{74.06\%} on UCF101, \textbf{51.63\%} on HMDB51 and \textbf{54.98\%} on PA-HMDB with $\epsilon$=$8$.
\section{Ablation Study}
Video-DPRP is also influenced by two major components in its algorithm: the projection dimensionality $k$ and the privacy budget $b$.\\
\underline{\textbf{Effect of varying the dimensionality $k$:}} For a fixed $\epsilon$$=$$8$, $\delta$$=$$10^{-4}$, and a privacy budget of $b$$=$$0.8$, we observed that increasing the dimensionality $k$  improves action recognition performance but results in a significant decrease in privacy, as shown in Table \ref{dimensionality}. This is because, the value of $k$, has a diminishing effect on the noise scale, $\sigma_p$$=$ $1/\sqrt{k}$ and also on the $L_{2}$ sensitivity, $\theta$ $=$ $255/\sqrt{k}$. As a result, when $k$ increases, it substantially reduces the standard deviation value $\sigma_1$ in Theorem \ref{theo3} and $\sigma_2$ in Theorem \ref{theo4}, leading to a decrease in the amount of noise added for differential privacy. Selecting an optimal $k$ requires balancing performance gains with acceptable privacy levels for practical viability.
\\\underline{\textbf{Effect of varying the privacy budget $b$:}} Recall that $b$, represents the privacy budget allocated to make the random projection differentially private, while $1-b$, ensures the differential privacy of the covariance matrix (see Algorithm \ref{alg:example}). To understand the effect of varying $b$, we fixed $\epsilon$$=$$8$, $\delta$$=$$10^{-4}$ and $k$$=$$32\times32\times3$. Table \ref{budget} shows that increasing the privacy budget for random projection up to a value of \textbf{80\%}, results in a less noisy random projection. Consequently, there is an increase in action recognition performance but with a substantial decrease in privacy. This suggests that the random projection plays a more critical role compared to the covariance matrix, in Video-DPRP.
\section{Conclusion}
This paper introduces Video-DPRP, a differentially private approach for constructing visual privacy-preserved videos for Human Activity Recognition (HAR). Video-DPRP aims to bridge the gap between visual privacy and utility by providing strong privacy guarantees through the mathematical properties of differential privacy and random projection. Our evaluation across multiple datasets demonstrate that Video-DPRP achieves competitive performance in activity recognition while maintaining robust privacy preservation compared to current state-of-the-art techniques. 